\pdfoutput=1

\documentclass[11pt]{article}

\usepackage[final]{acl}

\usepackage{times}
\usepackage{latexsym}
\usepackage{makecell}
\usepackage{listings}
\usepackage{xcolor}

\usepackage[T1]{fontenc}

\usepackage[utf8]{inputenc}

\usepackage{microtype}

\usepackage{inconsolata}

\usepackage{graphicx}

\title{{LNE-Blocking: An Efficient Framework for Contamination Mitigation Evaluation on Large Language Models}}

\author{\textbf{Ruijie Hou}\textsuperscript{1}\footnotemark[1],
    \textbf{Yueyang Jiao}\textsuperscript{1}\footnotemark[1], \\
    \textbf{Hanxu Hu}\textsuperscript{3}\footnotemark[2],
    \textbf{Yingming Li}\textsuperscript{1}, 
    \textbf{Wai Lam}\textsuperscript{4}, 
    \textbf{Huajian Zhang}\textsuperscript{5},
    and \textbf{Hongyuan Lu}\textsuperscript{2}\footnotemark[2]\\
   \textsuperscript{1}Zhejiang University 
   \textsuperscript{2}FaceMind Corporation \\
   \textsuperscript{3}University of Zurich 
   \textsuperscript{4}The Chinese University of Hong Kong 
   \textsuperscript{5}Westlake University \\
    \texttt{ruijie.hou@zju.edu.cn}, 
    \texttt{hongyuanlu@outlook.com}
}

\newcommand{\rebuttal}[1]{\textcolor{black}{#1}}

\usepackage{hyperref}
\usepackage{url}
\usepackage{algorithm}
\usepackage{algpseudocode}
\usepackage{multirow} 
\usepackage{array}
\usepackage{graphicx} 
\usepackage{booktabs}
 \usepackage{amsmath}
 \usepackage{amssymb}

\begin{document}
\maketitle
\renewcommand{\thefootnote}{\fnsymbol{footnote}}
\footnotetext[1]{Both authors contributed equally to this work.}
\footnotetext[2]{Hongyuan Lu and Hanxu Hu are corresponding authors.}
\renewcommand{\thefootnote}{\arabic{footnote}}
\begin{abstract}
The problem of data contamination is now almost inevitable during the development of large language models  (LLMs), with the training data commonly integrating those evaluation benchmarks even unintentionally. This problem subsequently makes it hard to benchmark LLMs fairly. 
Instead of constructing contamination-free datasets (quite hard), we propose a novel framework, \textbf{LNE-Blocking}, to restore model performance prior to contamination on potentially leaked datasets. Our framework consists of two components: contamination detection and disruption operation. 
For the prompt, the framework first uses the contamination detection method, \textbf{LNE}, to assess the extent of contamination in the model. Based on this, it adjusts the intensity of the disruption operation, \textbf{Blocking}, to elicit non-memorized responses from the model. Our framework is the first to efficiently restore the model's greedy decoding performance. This comes with a strong performance on multiple datasets with potential leakage risks, and it consistently achieves stable recovery results across different models and varying levels of data contamination. We release the code at \url{https://github.com/RuijieH/LNE-Blocking} to facilitate research.
\end{abstract}
\section{Introduction}
In the era of fierce development with large language models  (LLMs), it has been a popular research topic across many areas such as chain-of-thought reasoning \citep{wang-etal-2023-towards,10.5555/3600270.3602070}, machine translation \citep{2023arXiv230506575L,zhu-etal-2024-multilingual}, code generation \citep{li-etal-2023-codeie,zhang-etal-2023-self}, and even spatial reasoning \citep{hu2024chainofsymbol}. 
Despite the fact that LLMs are usually strong on many tasks, Natural Language Processing  (NLP) practitioners secretly face a common problem called data contamination
when conducting NLP research and engineering which frequently relies on benchmark data that could be potentially contaminated in LLMs.
\par
\begin{figure}[t]
\centering
  \includegraphics[width=0.95\columnwidth]{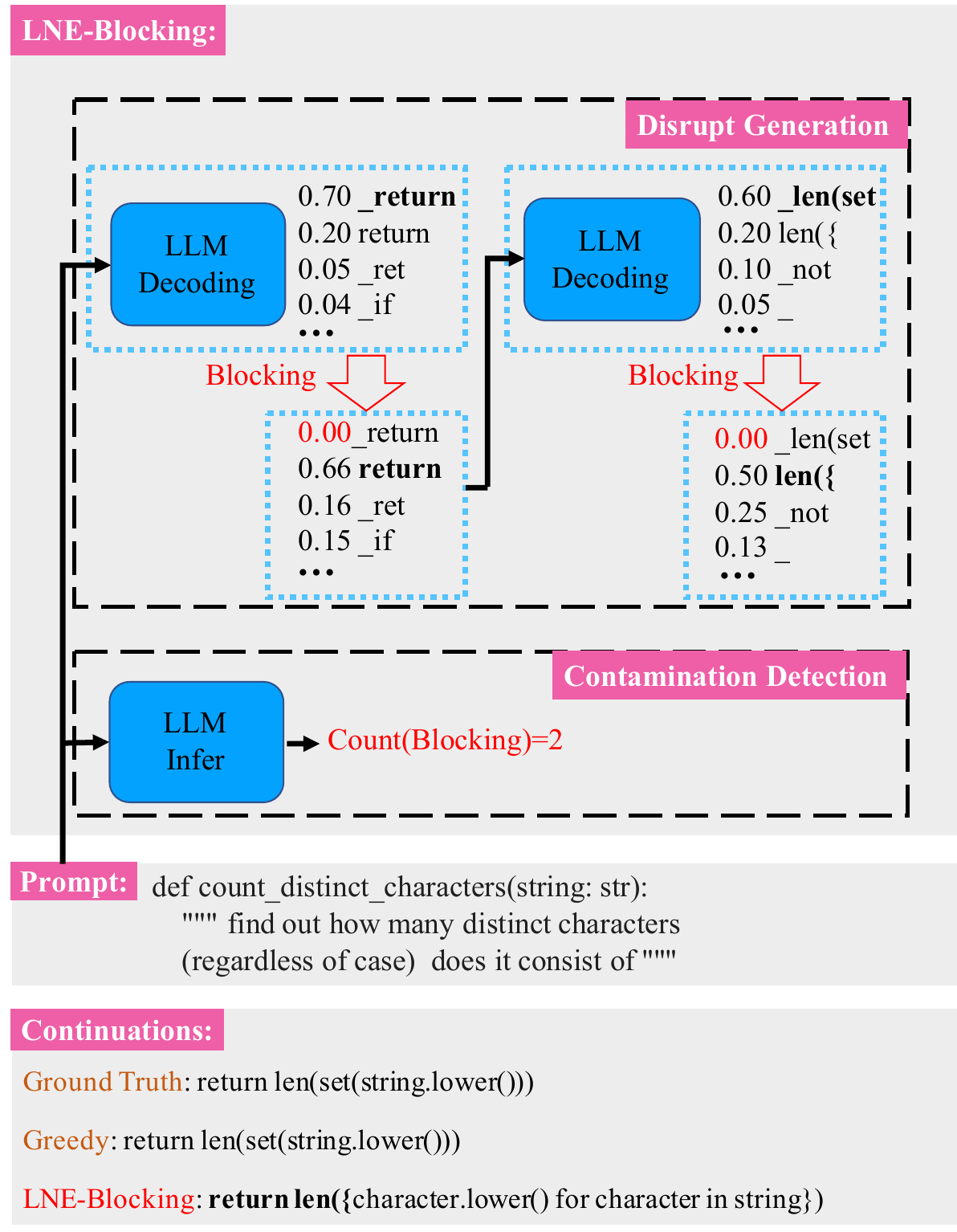}
  \caption{Illustration of our proposed LNE-Blocking framework. The whole framework decouples contamination mitigation evaluation into two components: contamination detection and disruption operations. The contamination detection step determines the number of Blocking operations to be performed, which are then executed to guide the model in generating a non-memorized response for the current sample.}
  
  \label{fig:lneblocking}
\end{figure}
Data contamination, also called data leakage, occurs when the test data is inadvertently included in the model’s training data \citep{magar-schwartz-2022-data,golchin2024time}. This causes the model to perform exceptionally well on the leaked test data. Due to the immense scale and diverse origins of the pre-trained datasets used for LLMs, even when developers do not intentionally introduce contamination, LLMs are more vulnerable to data contamination.
\par
As a result, preventing benchmark data contamination in LLMs becomes highly challenging. This prevents NLP developers and researchers from honestly judging the LLMs. To conduct the contamination-free evaluation of models, prior works \citep{cleaneval,li-etal-2024-gsm,zhang2024careful} have assessed model performance by creating or reconstructing new datasets that are free from leakage. However, these approaches are associated with significant labor costs and do not completely eliminate the risk of these datasets being inadvertently leaked in newly released models. One less studied problem, defined as \textbf{contamination mitigation evaluation}, is conducting the contamination-free evaluation of models on datasets already at risk of leakage.

For contamination mitigation evaluation,  TED~\citep{dong-etal-2024-generalization} spends significant time sampling and generating multiple responses on existing benchmarks, then removes similar responses to assess the model's genuine performance under sampling methods. 
In contrast, this paper proposes the LNE-Blocking framework, which adaptively restores the model's true performance under varying contamination levels. It operates online during generation, without relying on sampled responses, and directly assesses the model's genuine performance under greedy decoding.
\par 
Specifically, the LNE-Blocking framework explicitly decouples contamination mitigation into contamination detection and disruption operations, as shown in Figure \ref{fig:lneblocking}. The contamination detection strategy, LNE~(\textbf{L}ength \textbf{N}ormalized \textbf{E}ntropy), determines the degree of contamination based on the model's output, while the disruption operation, Blocking, intervenes in the original generation process by suppressing the token with the highest response probability during decoding. Overall, the framework adjusts the frequency of disruption operations based on the model's contamination level, prompting the model to generate non-memorized content for the current sample. To the best of our knowledge, we are the first to propose a method for assessing the model's genuine performance under greedy decoding. Additionally, experiments demonstrate that our approach is highly robust across different tasks and models, even with varying levels of contamination.

\par
To this end, we make three key contributions:
\begin{itemize}
\setlength\itemsep{0em}
    \item We propose the LNE-Blocking framework, which decouples contamination mitigation into contamination detection and disruption operations, enabling contamination-free model evaluation without relying on sampling methods.
    
    \item For the contamination mitigation evaluation task, we are the first to assess the model's genuine performance under greedy decoding, addressing a key gap in current research.

    \item Extensive experiments demonstrate that the proposed approach is highly robust across a wide range of tasks and LLMs.
    
\end{itemize}

\section{Motivation}
\begin{figure}
\centering
  \includegraphics[width=\columnwidth]{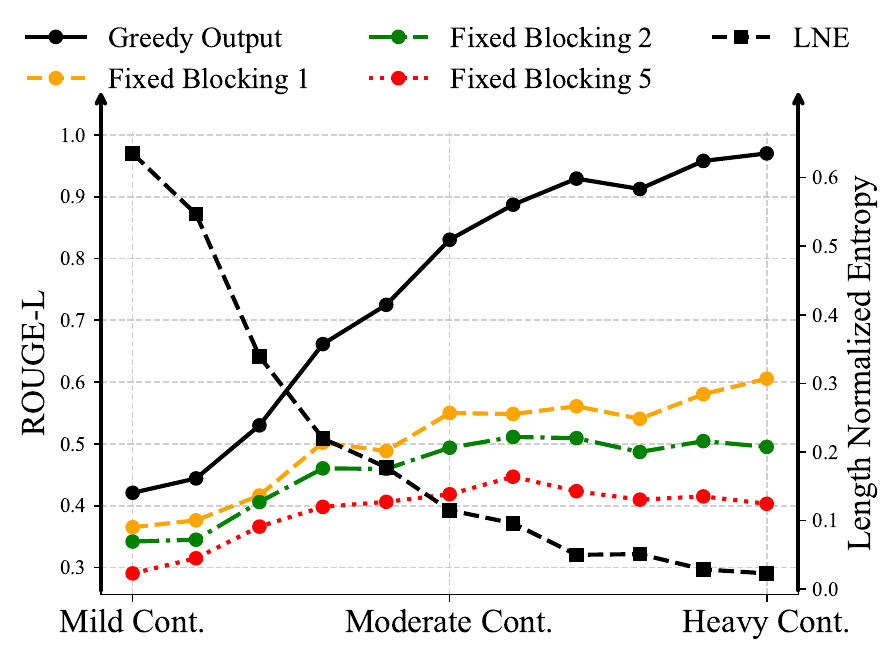}
  \caption{An example showing the changes in Length-Normalized Entropy (LNE), and the impact of the Blocking operations, on model memory phenomena (ROUGE-L), in models with varying levels of contamination, where "Cont." is an abbreviation of contamination. Fixed Blocking 1, 2, and 5 refer to the fixed number of Blocking operations applied for any given prompt.}
  
  \label{fig:motivation}
\end{figure}

Contaminated models often exhibit a high lexical overlap between their output and the ground truth, which is indicative of memory phenomena \citep{magar-schwartz-2022-data}. As shown in Figure \ref{fig:motivation}, as the degree of contamination increases, the overlap between the output generated by greedy decoding and the ground truth significantly rises. This behaviour reflects the model's tendency to memorize the training data rather than generalizing it. 
One solution \citep{dong-etal-2024-generalization} to assess the genuine performance of such contaminated models involves generating diverse outputs through multiple sampling and filtering out similar samples. Then, they expect to derive non-memorized answers that stem from the model's generalization abilities, rather than its memorized knowledge.
\par
However, answers based on memory phenomena tend to have a very high likelihood within the decoding process, meaning that obtaining non-memorized, generalized answers requires a large number of samplings. 
TED \citep{dong-etal-2024-generalization} has found that at least 50 samples are necessary to achieve satisfactory performance estimates. This process is both highly random and time-consuming.
Nevertheless, the inherent randomness of the sampling process presents a fundamental limitation: it makes consistent generation of non-memorized answers across models with varying contamination levels challenging, even when employing the recommended 50 sampling attempts. Particularly for heavily contaminated models, this approach fails because memorized outputs dominate the sampling distribution, and the probability of obtaining sufficient non-memorized samples is critically low for reliable evaluation. This fundamental limitation is empirically demonstrated in our subsequent experiments in section \ref{logistic_reson}. 
\par
To address this limitation, a controlled generation strategy is required to elicit non-memorized responses from the model.
When interrupted during question answering, the human could discard rote-memorized responses and reformulate answers based on a deeper conceptual understanding of the question. Inspired by this, we propose the \textbf{Blocking} operation, a technique that suppresses the generation of the highest-probability tokens during decoding, to simulate the interruption mechanism observed in human response patterns. It can be controllably triggered online during decoding, effectively overcoming the inherent randomness and time-intensive nature of the sampling method.

Blocking reduces the reliance on memorized content and encourages the model to generate non-memorized answers, as demonstrated by the reduction in ROUGE-L \citep{lin-2004-rouge} similarity between the output after applying the Blocking operation and the memorized output (denoted as Greedy Output), as in Figure \ref{fig:motivation}. Blocking enables the model to produce more diverse and generalized outputs, without sacrificing quality.
For example, the output of the contaminated model is \textit{return len(set(string.lower()))}. If the model is good at coding, after applying the Blocking operation, the model could generate an equivalent piece of code using its generalization ability, such as \textit{return len(\{character.lower() for character in string\})}\footnote{We provide additional examples in the Appendix \ref{sec:cases}.}.
\par
Additionally, as shown in Figure \ref{fig:motivation}, models with different contamination levels exhibit varying degrees of memorization, and the impact of the Blocking operation varies accordingly. 
To effectively disrupt memorized responses, models with varying contamination levels require different intensities of Blocking. In particular, as the level of contamination increases, the intensity of the Blocking must be increased to counteract the effects of memory.
\par
A natural idea, then, is to explicitly combine the contamination detection strategy and the Blocking operation into a unified framework, \textbf{LNE-Blocking}. In this framework, the first step is to detect the degree of contamination, and then use this information to adjust the intensity of Blocking, ensuring that memorization is disrupted without negatively affecting performance. This framework allows for a more targeted and adaptive application of Blocking across different contamination levels. 
And, we propose using \textbf{L}ength \textbf{N}ormalized \textbf{E}ntropy (\textbf{LNE}) as the contamination detection strategy. A heavily contaminated model will exhibit greater certainty in its token predictions. 
As a result, the entropy in the probability distribution at each decoding position becomes lower as the model grows more confident in generating memorized content. As shown in Figure \ref{fig:motivation}, with increased contamination, the generated text becomes closer to the ground truth, leading to a corresponding decrease in LNE.
\par

\section{Related Work}
\paragraph{Data Contamination Detection}
The issue of data contamination in large language models (LLMs) gained attention in the context of GPT-3 \citep{gpt3}, where the vast pre-training corpus inevitably overlapped with evaluation benchmarks. Meanwhile, as models are iteratively improved by using data coming from users, they overlook the problem of indirect data leakage \citep{balloccu-etal-2024-leak}. Following this, some work \citep{harmofcot1,harmofcot2,harmofcot3,harmofcot4} exposed the serious consequences of data contamination and urged attention to this problem. 
To address this, Min-k\% Prob \citep{mink} calculates the average of the smallest k\% probabilities of generated tokens and flags potential contamination if this average exceeds a certain threshold. Similarly, perplexity\citep{ppl} is also used to detect contamination, assuming that leaked data tends to produce lower perplexity scores.

\paragraph{Contamination Free Evaluation} 
\rebuttal{To evaluate LLMs in the context of potential data contamination, several methodologies generate new datasets that do not overlap with the model's training data, adopting a dataset-centric perspective. GSM-Plus\citep{li-etal-2024-gsm} ensures that benchmark data is absent from the model's training set by reconstructing the original GSM8k dataset \citep{cobbe2021gsm8k} through the introduction of perturbations. GSM1k\citep{zhang2024careful} involves the creation of a completely new dataset from scratch. It remains private and is only made publicly available at a future point in time. 
CleanEval\citep{cleaneval} proposed paraphrasing contaminated datasets using LLMs for evaluation. However, these approaches are associated with significant labor costs and do not fully eliminate the risk of these datasets being inadvertently leaked in newly released models.
}

One less explored but important problem is how to quantitatively assess the performance of models using datasets that have already been leaked. 
Some methods \citep{dynamic-eval-1, dynamic-eval-2, dynamic-eval-3} employ external LLMs as examiners to evaluate the performance of the target LLM. These frameworks typically necessitate assessing a broad spectrum of dimensions for each input in order to comprehensively measure the model’s understanding, which often incurs substantial computational and implementation overhead. Unlike approaches that rely on external LLMs, TED \citep{dong-etal-2024-generalization} filters non-memorized samples through multiple sampling rounds, using these samples for contamination mitigation evaluation. However, the sampling process is both highly random and time-consuming. \rebuttal{LNE-Blocking can be controllably triggered online during the decoding process, effectively overcoming the inherent randomness and time-intensive nature of the sampling method.}
\section{Methodology}
The section discusses how to utilize LNE for assessing the degree of contamination, how to employ Blocking for disruption, and how to integrate both methods to construct a framework for contamination mitigation evaluation.
\par
\subsection{LNE for Assessing the Degree of Contamination}

Given a prompt $x$ and a language model $M$, our goal is to assess the degree of contamination of $M$ with respect to this prompt. First, we perform a greedy decoding inference to obtain $y^\text{greedy}$. The greedy decoding process can be expressed as:
\begin{align}
\label{equ:gene_greedy}
y^\text{greedy}_i 
& =\arg\max_{{j \in V}}({M}(x,y_{1:i-1}^\text{greedy}))
\end{align}
where $y_i^{\text{greedy}}$ represents the token at position $i$ in the generated sequence, and $M(x, y_{1:i-1}^{\text{greedy}})$ denotes the logits output of the model $M$, which is a vector of dimension $\mathbb{R}^V$, with $V$ representing the size of the vocabulary. The token $y_i^{\text{greedy}}$ is selected by identifying the index $j$ that maximizes the corresponding logit value.

Then, based on the probability distribution at each position during the inference that generates $y^\text{greedy}$ with length $N$, we calculate the Length Normalized Entropy (LNE) as:
\begin{small}
\begin{equation}
        \begin{aligned}
            \mathrm{LNE}(M, x) &= \frac{1}{N} \sum_{i=1}^N H\left( y_i | M, x, y_{1:i-1}^{\mathrm{greedy}} \right)  \\ 
            &= -\frac{1}{N} \sum_{i=1}^N \sum_j^{V}  p\left (y_i=j\right) \log p\left (y_i=j\right)
        \end{aligned}
        \label{equ:lne}
\end{equation}
\end{small}
where $H(y_i | M, x, y_{1:i-1}^{\mathrm{greedy}})$ represents the entropy at the $i$-th position during the greedy decoding process,  $p\left (y_i=j\right)$ denotes the probability of the model generating the $j$-th token from the vocabulary Vocab at the $i$-th position, with $V$ representing the size of the token vocabulary. 
\par
\par
As the degree of contamination of the model increases, the $\mathrm{LNE}(M, x)$ decreases. We normalize it to $ \overline{\mathrm{LNE}(M, x)} $,  as shown in Equation~\eqref{equ:normlne}, so that it is proportional to the degree of contamination, with the normalized value\footnote{We divide $\mathrm{LNE}(M, x)$ by 2 primarily because we observed that, for different models, the range of LNE generally falls between 0 and 2.} lying within the range of 0 to 1.
\begin{equation}
    \label{equ:normlne}
    \overline{\mathrm{LNE}(M, x)} = 1 - \frac{\mathrm{LNE}(M, x)}{2}
\end{equation}
\subsection{Blocking for Disrupting Generation}

To disrupt the generation of LLMs, we propose the blocking operation during the decoding process, which suppresses the token with the highest probability at a certain position during decoding.
\par
Specifically, for a model $ M $ and a prompt $ x $, when the Blocking operation is applied to the $ i $-th position during the decoding process, the resulting response is defined as:
\begin{equation}
\label{equ:block_resp}
     y^{\text{Blocking}(M,x,(i))} = ( y_{1:i-1}^\text{greedy},  y_i^\text{block}, y_{i+1:l}^\text{greedy} )
\end{equation}
where $l$ is the length of the resulting response. 
During the process, each position before the $i$-th applies greedy decoding using Equation~\eqref{equ:gene_greedy}. At the 
$i$-th position, the token with the highest probability from the distribution, modified by the Blocking operation, is selected for generation, as follows:
\begin{equation}
\label{equ:decode_block}
y_i^\text{block} =argmax (M_\text{block} (x,y_{1:i-1}^\text{greedy}))
\end{equation}
To obtain $M_\text{block} (x,y_{1:i-1}^\text{greedy})$, the logits output with the maximum value suppressed, the process involves first identifying the index of the maximum value and then suppressing it:
\begin{align}
    \scriptstyle M_\text{block}(x, y_{1:i-1}^\text{greedy}) \leftarrow M(x, y_{1:i-1}^\text{greedy}), \notag \\
    \scriptstyle M_\text{block}(x, y_{1:i-1}^\text{greedy})[\arg\max_{j \in V}({M_\text{block}(x, y_{1:i-1}^\text{greedy})})] &\leftarrow -\infty
\end{align}
After the generation of the $i$-th token, the subsequent tokens are still generated using greedy decoding:
\begin{align}
\label{equ:decode_after_greedy}
y_{i+1}^\text{greedy} &= \arg\max_{j \in V} ({M}(x,(y_{1:i-1}^\text{greedy},y_i^\text{block}))) 
\end{align}
\par
\subsection{LNE-Blocking for Contamination Mitigation Evaluation}
For models with different contamination levels, varying Blocking intensities are required during generation to disrupt the model's memorization, reflecting the original capabilities of the models. This section will sequentially explain how to control the Blocking intensity and how to determine it based on the level of contamination.
\par
\subsubsection{Multi Blocking operations for Disrupting Memorization}
For highly contaminated data, only performing the Blocking operation once during generation is not sufficient to disrupt its memorization, requiring more Blocking operations to increase disruption intensity. Meanwhile, Blocking is applied earlier in the generation process to interrupt the generation of memorized answer tokens as early as possible, ensuring that the final response is more likely to differ significantly from the standard memorized answer. It provides the model with sufficient room to adjust its response logic effectively. Another reason is that a study \citep{NEURIPS2024_7a8e7fd2} has shown that performing the sampling operation at the beginning of generation can trigger the model's chain-of-thought (COT) reasoning, reflecting its generalization ability.

Specifically, for a model $ M $ and a prompt $ x $, when the Blocking operation is applied $ n $ times starting from the first token during decoding, the resulting response is defined as:
\begin{equation}
\label{equ:gene_block_n_times}
     y^{\text{Blocking}(M,x,(1,2,...n))} = (y_{1:n}^\text{block}, y_{n+1:l}^\text{greedy} )
\end{equation}
First, for $i \leq n$, we define $y_i^{\text{block}}$ as:
\begin{equation}
    y_i^{\text{block}} = \arg\max_{j \in V} M_{\text{block}}(x, y_{1:i-1}^{\text{block}}).
\end{equation}
Then, for $i > n$, the remaining tokens are generated as:
\begin{equation}
    y_i^{\text{greedy}} = \arg\max_{j \in V} M_{\text{block}}(x, (y_{1:n}^{\text{block}}, y_{n:i-1}^{\text{greedy}})).
\end{equation}

\par

\subsubsection{Determine the disrupting intensity based on the LNE}

The Blocking intensity is determined based on the contamination level, detected by $\overline{\mathrm{LNE}(M, x)}$. 
Given a prompt $x$ and model $M$, $\overline{\mathrm{LNE}(M, x)}$ is first obtained through greedy decoding using Equation~\eqref{equ:normlne}, and then the Blocking intensity corresponding to both the prompt and the model is controlled by determining the number of Blocking operations using $\overline{\mathrm{LNE}(M, x)}$ and Threshold\_Task jointly.
\par
Specifically, the number of Blocking operations is defined as:
\begin{small}
\begin{equation}
\label{equ:count_block}
Cnt (M, x)= \mathrm{round}(\overline{\mathrm{LNE}(M, x)}*Threshold\_Task)
\end{equation}
\end{small}
where $Threshold\_Task$ is a hyperparameter dependent on the specified task. On each task, each data-model pair undergoes greedy generation once, and the corresponding number of Blocking operations can be determined using Equation~\eqref{equ:count_block}.
\par
Since the range of $\overline{\mathrm{LNE}(M, x)}$ is between 0 and 1, for heavily polluted (memorized) samples, $\overline{\mathrm{LNE}(M, x)}$ can reach its maximum value of 1. In this case, the corresponding number of Blocking operations, $Cnt (M, x)$, also attains its maximum value, Threshold\_Task\footnote{We detail in the Appendix \ref{sec:threshold} how the hyperparameter Threshold\_Task is determined for different tasks.}. Therefore, Threshold\_Task acquires the following practical interpretation: it represents the maximum number of times a sample can be blocked under the given task.
Notably, this hyperparameter depends on the evaluation task but is independent of the model. Therefore, when the evaluation task is known but the contamination level of the model is unknown, this threshold can still be determined.
\subsubsection{Contamination Mitigation Evaluation}
After performing greedy generation once to obtain $\overline{\mathrm{LNE}(M, x)}$ using Equation~\eqref{equ:normlne}, we calculate the number of Blocking operations, $Cnt (M, x)$, using Equation~\eqref{equ:count_block}. Then, we perform the Blocking operation $Cnt (M, x)$ times to disrupt memorization. The resulting output is defined as:
\begin{equation}
\label{equ:lneblocking}
    y^{\text{LNE-Blocking}}=y^{\text{Blocking}(M,x,(1,2,...Cnt (M, x)))}
\end{equation}

Finally, for an evaluation metric $\mathcal{E}$, $\mathcal{E} (y^\text{LNE-Blocking})$ is used instead of $\mathcal{E} (y^\text{greedy})$ to evaluate the model's performance after the contamination mitigation under greedy decoding.
\par
The LNE-Blocking pseudocode for contamination mitigation evaluation is shown in Algorithm \ref{alg:LNE_block}.
\begin{algorithm}
\caption{The pseudocode of LNE-Blocking }
\label{alg:LNE_block}
\begin{algorithmic}[1] 
\Require LLM $M$, the prompt of test data $x$, evaluation metric $\mathcal{E}$, and hyper-parameter $Threshold\_Task$.
\Ensure Genuine Performance under greedy decoding $ep$.
\State Obtain $y^\text{greedy}$ from $M$ with the prompt $x$ via Equation~\eqref{equ:gene_greedy}.
\State Get the $\overline{\mathrm{LNE}(M, x)}$ via Equation~\eqref{equ:normlne}.
\State Determine the execution count of the Blocking operation, \(Cnt(M, x)\), use Equation~\eqref{equ:count_block}.

\State Obtain $y^\text{LNE-Blcoking}$ via Equation~\eqref{equ:lneblocking}.
\State Obtain $ep$ based on $\mathcal{E} (y^\text{LNE-Blocking})$.
\State \Return $ep$.
\end{algorithmic}
\end{algorithm}
\section{Experimental Setup}
\subsection{Dataset}
\paragraph{HumanEval \citep{chen2021humaneval}:}
The HumanEval dataset released by OpenAI includes 164 programming problems with a function signature, docstring, body, and several unit tests, all handwritten to ensure exclusion from the training set of code generation models. And the initial publications\citep{touvron2023llama2,roziere2023codellama,nijkamp2022codegen,dubey2024llama3} of the models - Llama 2, CodeLlama, CodeGen and Llama 3.1, employ the HumanEval dataset as a benchmark for evaluating code generation performance. We assumed that these models have not been contaminated by the test set of the HumanEval dataset.
\paragraph{GSM8K \citep{cobbe2021gsm8k}:}
GSM8K  (Grade School Math 8K) is a dataset of 8.5K high-quality linguistically diverse grade school math word problems. The dataset was created to support the task of question answering on basic mathematical problems that require multi-step reasoning.  The initial publications \citep{touvron2023llama2,dubey2024llama3} of the models, Llama 2, Llama 3.1, employ the GSM8K dataset as a benchmark for evaluating their arithmetic reasoning capacity. 
\paragraph{GSM-Plus~\citep{li-etal-2024-gsm}:}
It is an augmented version of GSM8K with various mathematical perturbations, including numerical variation, arithmetic variation, problem understanding challenges, distractor insertion, and critical thinking tasks. This dataset was released in January 2024, which is after the release of Llama 2. Given the randomness of these perturbations and the innovative nature of the techniques employed, it is highly likely that the original uncontaminated version of Llama 2 was not exposed to contamination during its training.

\par
\subsection{Models}
For the code generation task, we chose four models, Llama 2, CodeLlama, CodeGen and Llama 3.1, each trained for 20 epochs to produce 20 LoRA weights corresponding to different levels of contamination.
For Llama 2, CodeLlama, and CodeGen, the contaminated models were directly simulated using the LoRA weights provided by TED\citep{dong-etal-2024-generalization}, which simulate data contamination by training LLMs using benchmark data, \rebuttal{mixing the HumanEval test set and StarCoder data\citep{li2023starcoder} at 1:1,000 ratio}. For the recent model, Llama 3.1, we used its base version and employed a continued pretraining approach using the test set of the HumanEval dataset to simulate contamination as TED\citep{dong-etal-2024-generalization}.

\par
For the arithmetic reasoning task, we utilize the base versions of the Llama 2 and Llama 3.1 models and adopt a continued pretraining methodology using the test set of the GSM8K dataset to emulate contamination, executing training for 20 epochs. To simulate more realistic contamination scenarios, we apply the same continued pretraining strategy to Llama 2 using the GSM-Plus dataset.

For ease of analysis, we defined the first third of the 20-epoch contamination as mildly contaminated (Mild Cont.), the middle third as moderately contaminated (Moderate Cont.), and the final third as heavily contaminated (Heavy Cont.). The training was conducted on a single 4090 GPU using the LLaMA-Factory framework \citep{zheng2024llamafactory}, with a learning rate of 1e-4. And the training time for the code generation and arithmetic reasoning tasks was 2 hours and 20 hours, respectively.
\subsection{Evaluation Metrics}

\par
For contamination mitigation evaluation, we measure the model's performance after contamination has been mitigated. Specifically, the performance metrics used for code generation and arithmetic reasoning tasks are \textbf{Pass@1} and exact match \textbf{Accuracy} \citep{dong-etal-2024-generalization}, respectively. Additionally, we introduce a novel metric, \textbf{P}erformance \textbf{G}ap (\textbf{PG}), defined as:
\begin{equation}
    \label{metric:pg}
    \text{PG} = abs (\mathcal{E} (Y^{eva}_{M}) - \mathcal{E} (Y_{M_{origin}}))
\end{equation}
where $Y^{eva}_{M}$ represents the output of the model after contamination mitigation on the entire test dataset, and $Y_{M_{origin}}$ represents the output of the corresponding uncontaminated model on the entire test dataset. For LNE-Blocking, $Y^{eva}_{M}$ and $Y_{M_{origin}}$ correspond to $Y^\text{LNE-Blocking}_{M}$ and $Y^\text{greedy}_{M_{origin}}$, while for TED, they correspond to $Y^{TED}_{M}$ and $Y^{sampling}_{M_{origin}}$.

PG quantifies how closely the performance of the model after mitigation matches the original uncontaminated model. A smaller PG value indicates that the contamination mitigation strategy is better.
\section{Results}

\subsection{Contamination Mitigation Evaluation}
\begin{table*}[!htbp] 
\setlength\tabcolsep{6pt}
\small
\centering
\caption{Contamination mitigation evaluation on Code Generation, where values outside the parentheses represent model performance, Pass@1, while those inside the parentheses represent the PG metric. \textbf{Bold} indicates the strategy with the best performance at the current contamination level, and \underline{underline} highlights our proposed strategy, which significantly outperforms others under the current contamination level.}
\label{tab:eva_code}

\resizebox{0.95\textwidth}{!}{%

\begin{tabular}{lllllll}
\toprule Model & Strategy & Uncontaminated & Mild Cont. & Moderate Cont. & Heavy Cont. & Average \\
\midrule \multirow{4}{*}{ CodeGen-6B } & Sampling & 0.122 & 0.246 & 0.714 & 0.836 & 0.577 \\
 & Greedy & 0.165 & 0.317 & 0.819 & 0.913 & 0.653 \\
 & TED & 0.106  (\textbf{0.016}) & 0.148  (\textbf{0.036}) & 0.234   (0.112) & 0.211  (0.072) & 0.188   (0.072) \\
 & LNE-Blocking & 0.073  (0.091) & 0.094  (0.071) & 0.113  (\textbf{0.052}) & 0.117  (\textbf{0.037}) & 0.108  (\textbf{0.056}) \\
\midrule \multirow{4}{*}{ Llama 2-7B } & Sampling & 0.111 & 0.243 & 0.659 & 0.798 & 0.556 \\
 & Greedy & 0.128 & 0.289 & 0.742 & 0.861 & 0.609 \\
 & TED & 0.095  (\textbf{0.016}) & 0.118   (0.017) & 0.128   (\textbf{0.021}) & 0.114   (0.036) & 0.114   (\textbf{0.024}) \\
 & LNE-Blocking & 0.098  (0.030) & 0.144  (\textbf{0.016}) & 0.134  (0.037) & 0.128  (\textbf{0.018}) & 0.132  (0.025) \\
\midrule \multirow{4}{*}{ CodeLlama-7B } & Sampling & 0.218 & 0.382 & 0.700 & 0.808 & 0.613 \\
 & Greedy & 0.311 & 0.447 & 0.784 & 0.870 & 0.682 \\
 & TED & 0.205   (\textbf{0.013}) & 0.318   (0.099) & 0.392  (0.174) & 0.375   (0.137) & 0.345   (0.129) \\
 & LNE-Blocking & 0.268   (0.043) & 0.307  (\textbf{0.033}) & 0.282   (\underline{\textbf{0.032}}) & 0.271   (\underline{\textbf{0.045}}) & 0.283  (\underline{\textbf{0.037}}) \\
\midrule \multirow{4}{*}{ Llama 3.1-8B } & Sampling & 0.329 & 0.474 & 0.879 & 0.947 & 0.739 \\
 & Greedy & 0.348 & 0.524 & 0.893 & 0.936 & 0.758 \\
 & TED & 0.306  (\textbf{0.023}) & 0.397  (0.084) & 0.257  (0.083) & 0.176  (0.169) & 0.273  (0.101) \\
 & LNE-Blocking & 0.293  (0.055) & 0.356  (\textbf{0.061}) & 0.364  (\textbf{0.038}) & 0.305  (\underline{\textbf{0.067}}) & 0.333  (\textbf{0.054}) \\
\bottomrule
\end{tabular}

}
\end{table*}
In this section, we evaluate the models' performance after applying the LNE-Blocking strategy to mitigate contamination. We also employ PG to evaluate the effectiveness of LNE-Blocking and compare these with TED \citep{dong-etal-2024-generalization}, a method for contamination mitigation evaluation based on sampling. The
definition of TED is illustrated in Appendix \ref{sec:ted}. 
\par

Contamination mitigation evaluation is conducted on two tasks: code generation and arithmetic reasoning. In Appendix \ref{sec:summarization}, we validate the effectiveness of our approach on the task of Summarization.
Additionally, relying solely on the declarations from model developers may not guarantee that the origin model has not been contaminated by the test dataset. To address this, we also utilized a recently released dataset, GSM-Plus, to validate the effectiveness of the contamination mitigation strategy on arithmetic reasoning task. Moreover, Appendix \ref{sec:phi} presents an analysis of our method's performance on smaller, domain-specific models.

\subsubsection{Code Generation}
For this task, we use the test set of HumanEval as the benchmark and set the $Threshold\_Task$ to 4. For the TED method, the edit distance threshold is set to 2, following \citep{dong-etal-2024-generalization}.
\par

As shown in Table \ref{tab:eva_code}, the PG metric of LNE-Blocking after contamination mitigation remains small, denoting that the LNE-Blocking strategy enables models to achieve relatively stable performance restoration across different models and contamination levels after contamination mitigation under greedy decoding. Additionally, the average PG metric of LNE-Blocking is small, denoting that after applying a contamination mitigation strategy to contaminated models, the evaluation performance approaches the performance of the original uncontaminated models. 
\par

In contrast, the PG metric of TED diverges from the original model as contamination deepens, indicating insufficient stability in restoration. 
Particularly in the heavily contaminated models CodeLlama and Llama 3.1, our method significantly outperforms TED
. 
This is mainly due to the randomness of sampling, which causes the TED method to fail on heavily contaminated models, while the Blocking operation avoids it by controlling the triggering of sampling.
\par

Furthermore, for models, CodeGen and Llama 2, with lower contamination, the LNE-Blocking strategy under-performs compared to TED. This may be because, at lower contamination levels, multiple samplings yield more diverse results to reduce memorization. This suggests that the framework has room for improvement with a more fine-grained strategy for detecting contamination levels.
\par
\subsubsection{Arithmetic Reasoning}
\label{logistic_reson}
\begin{table*}[!ht]  
\setlength\tabcolsep{6pt}
    \centering
    \small
\caption{Contamination mitigation evaluation on Arithmetic Reasoning datasets, GSM8K and GSM-Plus, where values outside the parentheses represent model accuracy, while those inside the parentheses represent the PG
metric.}
\label{tab:eva_logical}

\resizebox{0.95\textwidth}{!}{
\begin{tabular}{lllllll}
\toprule Model & Strategy & Uncontaminated & Mild Cont. & Moderate Cont. & Heavy Cont. & Average \\
\midrule 
\multicolumn{7}{c}{GSM8K Dataset} \\
\midrule \multirow{4}{*}{ Llama 2-7B } & Sampling & 0.221 & 0.380 & 0.715 & 0.874 & 0.627 \\
 & Greedy & 0.145 & 0.252 & 0.637 & 0.853 & 0.556 \\
 & TED & 0.217  (\textbf{0.005}) & 0.348   (0.126) & 0.278   (0.119) & 0.113   (0.162) & 0.232   (0.122) \\
 & LNE-Blocking & 0.133  (0.012) & 0.163  (\textbf{0.023}) & 0.224  (\textbf{0.079}) & 0.222  (\underline{\textbf{0.075}}) & 0.198  (\textbf{0.057}) \\
\midrule \multirow{4}{*}{ Llama 3.1-8B } & Sampling & 0.719 & 0.772 & 0.939 & 0.995 & 0.889 \\
 & Greedy & 0.555 & 0.592 & 0.889 & 0.993 & 0.807 \\
 & TED & 0.704  (\textbf{0.016}) & 0.723  (\textbf{0.019}) & 0.306  (0.414) & 0.050  (0.694) & 0.379  (0.346) \\
 & LNE-Blocking & 0.475  (0.080) & 0.429  ({0.126}) & 0.487  (\underline{\textbf{0.068}}) & 0.488  (\underline{\textbf{0.065}}) & 0.471  (\underline{\textbf{0.084}}) \\
\midrule 
\multicolumn{7}{c}{GSM-Plus Dataset} \\
\midrule 
\multirow{4}{*}{Llama 2-7B}
& Sampling & 0.151 & 0.274 & 0.683 & 0.752 & 0.542 \\
         & Greedy & 0.090  & 0.208  & 0.651 & 0.747 & 0.505  \\ 
          & TED & 0.152  (\textbf{0.001}) & 0.262   (0.111) & 0.305   (0.154) & 0.143   (\textbf{0.008}) & 0.235   (0.089) \\
         & LNE-Blocking & 0.096 (0.006) & 0.117 (\underline{\textbf{0.027}}) & 0.140 (\underline{\textbf{0.051}}) & 0.139 (0.049) & 0.130 (\textbf{0.040}) \\

\bottomrule
\end{tabular}

}

\end{table*}

For this task, we use the test set of the GSM8K and GSM-Plus dataset as the benchmark and set the $Threshold\_Task$ to 7. For the TED method, since previous research did not evaluate this task \citep{dong-etal-2024-generalization}, we conducted a search to identify the optimal threshold that is compatible with both Llama 2 and Llama 3.1, finding it to be 50.

\par

As shown in Table \ref{tab:eva_logical}, similar to the task of code generation, our method enables models to achieve relatively stable performance restoration across different models and contamination levels after contamination mitigation under greedy decoding. 

However, TED suffers from insufficient stability in restoration.
Particularly in the heavily contaminated models, our method significantly outperforms TED. It is worth noting that the PG metric increases dramatically to about 0.414 and 0.694 when applying TED to the mildly and heavily contaminated Llama 3.1, denoting it fails completely. This is also due to its sampling randomness, which prevents it from generating diverse answers when the probability of memorized answers is high. 

Meanwhile,  for Llama 2---which has lower
contamination risks on GSM-Plus---our method
achieves a performance restoration with a maximum Performance Gap (PG) of only 5\% across all contamination levels.
\begin{table*}[!htbp] 
\setlength\tabcolsep{9pt}
\caption{Ablation study of the components of LNE-Blocking, where values outside the parentheses represent model performance, Pass@1, while those inside the parentheses represent the PG metric. Perplexity is denoted as PPL, and Min-k\% Prob is denoted as Min-k, with their definitions provided in Appendix \ref{sec:detect_baseline}. \rebuttal{ Fixed Blocking 1, 2, and 5 refer to the fixed number of Blocking operations applied for any given prompt.}}
\footnotesize
\centering
\label{tab:ablation}
\resizebox{0.95\textwidth}{!}{
\begin{tabular}{lllllll}
\toprule Strategy & Uncontaminated &  Mild Cont. & Moderate Cont. & Heavy Cont. & Average\\
\midrule Greedy Decoding & 0.311 & 0.447 & 0.784 & 0.870 & 0.682\\
 Fixed Blocking 1 & 0.287 (\textbf{0.024}) & 0.364 (0.089) & 0.430 (0.119) & 0.413 (0.100) & 0.394 (0.097) \\
 Fixed Blocking 2 & 0.274 (0.037) & 0.360 (0.089) & 0.372 (0.070) & 0.351 (\textbf{0.033}) & 0.352 (0.062) \\
 Fixed Blocking 3 & 0.250 (0.061) & 0.329 (0.047) & 0.305 (0.037) & 0.288 (0.035) & 0.304 (0.041) \\
 Fixed Blocking 4 & 0.165 (0.146) & 0.266 (0.065) & 0.274 (0.040) & 0.271 (0.043) & 0.261 (0.057) \\
 PPL-Blocking & 0.262 (0.049) & 0.313 (0.043) & 0.280 (0.037) & 0.274 (0.047) & 0.284 (0.042) \\
  Min-k-Blocking & 0.274 (0.037) & 0.339 (0.065) & 0.274 (0.051) & 0.270 (0.043) & 0.290 (0.052) \\
 LNE-Blocking & 0.268 (0.043) & 0.307 (\textbf{0.033}) & 0.282 (\textbf{0.032}) & 0.271 (0.045) & 0.283 (\textbf{0.037}) \\
\bottomrule
\end{tabular}

}
\end{table*}

\begin{table*}[!ht]
    \setlength\tabcolsep{9pt}
    \centering
    \caption{Coherence Measurement for Arithmetic Reasoning Using Llama-3.1, where PPL refers to Perplexity, GPTS to GPT Score, and HumanE to Human Evaluation. And LNE-Blocking is denoted as LB.}
    \label{tab:coherencellama22}
    \resizebox{0.95\textwidth}{!}{
    \begin{tabular}{llllllll}
     \toprule
        Metric & Ground Truth & \makecell[l]{Uncont.\\pre-LB} & \makecell[l]{Uncont.\\post-LB} & \makecell[l]{Mild Cont.\\post-LB} & \makecell[l]{Moderate Cont.\\post-LB} & \makecell[l]{Heavy Cont.\\post-LB} & \makecell[l]{Average\\post-LB} \\  \midrule
        PPL & 11.189 & 10.260 & 11.007 & 11.176 & 12.354 & 12.992 & 11.977 \\ 
        GPTS & 8.408 & 8.106 & 7.503 & 7.434 & 6.902 & 6.803 & 7.123 \\ 
        CER & 100.0\% & 8.5\% & 11.0\% & 7.2\% & 29.1\% & 39.6\% & 22.3\% \\ \bottomrule
    \end{tabular}
    }
\end{table*}
\subsection{Ablation study}
In this section, we analyze the contribution of each component in LNE-Blocking and examine how text generation coherence changes before and after its application. And we analyze the selection of the hyperparameter Threshold\_Task in the Appendix \ref{sec:threshold}. 
\subsubsection{Effectiveness of LNE-Blocking Components}
As shown in Table \ref{tab:ablation},
when using a fixed number of Blocking operations to restore the performance of models with varying levels of contamination, the extent of performance restoration differs. With fewer Blocking operations, the performance of models with mild contamination can be well restored, but heavily contaminated models remain poorly restored. And, as the number of Blocking operations increases, models with more severe contamination are restored more effectively, while the restoration of less contaminated models becomes less optimal. This demonstrates the effectiveness of employing a contamination detection strategy to adjust the Blocking intensity according to the contamination level.
\par
When using other existing contamination detection methods, such as Perplexity and Min-k\% Prob, instead of LNE to adjust the Blocking intensity, their performance recovery is less effective than that achieved with LNE, as shown in Table \ref{tab:ablation}. This highlights that LNE is more suitable for adjusting Blocking intensity, possibly because LNE leverages more information from the entire distribution at each decoding position.
\subsubsection{Impact on Generation Coherence}
We conducted coherence evaluations for code generation tasks employing GPT-based scoring, Perplexity (PPL) metrics and Compilation Error Rate of the generated code. Specifically, the Compilation Error Rate is defined as the ratio of SyntaxError and IndentationError occurrences during runtime over the total number of samples for the code generation task. This metric could accurately reflect the consistency of the generated code content. Details of the remaining metrics are provided in the Appendix \ref{sec:coherence}. 

As shown in Table \ref{tab:coherencellama22}, the coherence metrics show that as the level of contamination increases, the coherence of the outputs after blocking slightly deteriorates. However, the changes remain at a relatively low level, indicating that the impact on coherence is minimal. While during this process, the model's accuracy after blocking significantly drops compared to its accuracy under contamination. Especially, when applying LNE-Blocking to mildly contaminated models, the blocked model’s generated outputs exhibit a lower compilation error rate (7.2\%) than the contaminated model (8.5\%). This strongly suggests that blocking does not disrupt the coherence of the generated content.
\section{Conclusion}
In this paper, we propose the LNE-Blocking framework to address the challenge of data contamination in large language models (LLMs). By decoupling detection and disruption, the framework first restores the model's greedy decoding performance after contamination mitigation. Through extensive experiments, we demonstrate that LNE-Blocking effectively restores model performance under greedy decoding, achieving robust and consistent promising results across diverse tasks and contamination levels. This work provides a practical and efficient solution for contamination mitigation, offering a new direction for fair benchmarking and reliable evaluation of LLMs.

\section{Limitations}
Our work has several limitations, which we aim to address in our future work:

First, the evaluation of our work is mainly focused on benchmarks for code generation and arithmetic reasoning. In the future, we will further validate our approaches on other benchmarks.

Second, considering the limitation of computational resources, we employ LoRA instead of full-parameter fine-tuning to simulate data contamination for LLMs. In future work, we plan to extend our setting to full-parameter fine-tuning.

Third, considering the issue of training cost, we currently simulate contamination through the method of continued pretraining. In real-world scenarios, a significant portion of contamination also arises from pretraining from scratch. To simulate this contamination, it might have to retrain an LLM from scratch using a large corpus that includes some test data. However, such a process would be prohibitively expensive.

\bibliography{custom}

\clearpage
\appendix
\section{\rebuttal{Formulas and Principles for Comparison Methods}}
\subsection{Data Contamination Detection}
\label{sec:detect_baseline}
  
Consider the output of the model's greedy decoding as $ y $. Below are the definitions of several contamination detection methods.

\textbf{Perplexity}: calculates the perplexity of the response generated by the model through greedy decoding, as:

\begin{equation}
    \label{equ:ppl}
\text{Perplexity} = \exp\left( -\frac{1}{N} \sum_{i=1}^{N} \log P(y_i ) \right)
\end{equation}
where N is the length of the greedy decoded output, $y$.
Lower perplexity indicates higher contamination levels.

\textbf{Mink\% Prob}: Computes the negative average log probability of the k\% least
probable tokens in the response generated by the model through greedy decoding, as:
\begin{equation}
    \label{equ:mink}
   \text{Min-k}(y) = -\frac{1}{E} \sum_{y_i \in \text{Min-k\%(y)}}\log P(y_i)
\end{equation}
where E is the size of the Min-K\%(y) set, and Min-K\%(y) is the set of  k\% least probable tokens in the response. Smaller values of Min-k\% Prob indicate higher contamination levels.

\subsection{Contamination Mitigation Evaluation}
\label{sec:ted}
TED: filters non-memorized samples through multiple sampling rounds, using these samples
for contamination mitigation evaluation. It is illustrated as:
\begin{equation}
    \label{equ:ted}
S_e=\left\{s \mid s \in S \wedge \mathrm{EditDist}\left(s, s^\text{greedy}\right)>\tau\right\}
\end{equation}
where $ S $ is the set of outputs sampled from LLM, $ s^\text{greedy} $ denotes the output generated by the LLM using greedy decoding, and $ \tau $ is a predefined threshold for the edit distance. The set $ S_e $ contains only those samples from $ S $ for which the edit distance to the greedy decoding output exceeds the threshold $ \tau $.

\section{Contamination Mitigation Evaluation on the Summarization Task}\label{sec:summarization}
To achieve a broader validation of the effectiveness of LNE-Blocking across  benchmarks, we conducted experiments on a summarization task using the ACLSum \citep{takeshita-etal-2024-aclsum} dataset. This dataset, released in 2024, is specifically designed for aspect-based summarization of scientific publications. The models evaluated were Qwen2.5-7B and Qwen2.5-14B, and we employed the ROUGE-L metric to measure the similarity between model-generated outputs and the reference summaries. The $Threshold\_Task$ was set to 30, and both contamination simulation and inference were conducted using a two-shot prompt.
\begin{table*}[!htbp]  
\setlength\tabcolsep{6pt}
\caption{Contamination mitigation evaluation on Summarization, where values outside the parentheses represent model performance, ROUGE-L, while those inside the parentheses represent the PG
metric based on ROUGE-L.}
    \centering
    \small

\label{tab:eva_sum}

\resizebox{\textwidth}{!}{
\begin{tabular}{lllllll}
\toprule Model & Strategy & Uncontaminated & Mild Cont. & Moderate Cont. & Heavy Cont. & Average \\
\midrule \multirow{2}{*}{ Qwen2.5-7B}  & Greedy & 0.145 & 0.220 & 0.248 & 0.253 & 0.230 \\ 
        & LNE-Blocking & 0.135 (0.011) & 0.175 (0.013) & 0.179 (0.011) & 0.152 (0.011) & 0.166 (0.011) \\ 
        \midrule
        \multirow{2}{*}{ Qwen2.5-14B} & Greedy & 0.139 & 0.226 & 0.265 & 0.304 & 0.250 \\ 
        & LNE-Blocking & 0.122 (0.017) & 0.175 (0.036) & 0.163 (0.024) & 0.143 (0.005) & 0.158 (0.023) \\ 

\bottomrule
\end{tabular}

}
\end{table*}
As shown in Table \ref{tab:eva_sum}, our strategy consistently and stably restores the genuine capabilities of models with varying levels of contamination for both Qwen2.5-7B and Qwen2.5-14B on the summarization task. These results suggest that our approach is effective not only in code generation and arithmetic reasoning but also in NLP tasks like summarization.
\begin{table*}[!htbp]  
\setlength\tabcolsep{6pt}
\centering
\caption{Contamination mitigation evaluation on HumanEval for Phi-1, where values outside the parentheses represent model performance, Pass@1, while those inside the parentheses represent the PG
metric.}
\label{tab:phi}
\resizebox{\textwidth}{!}{
\begin{tabular}{lllllll}
\toprule Model & Strategy & Uncontaminated & Mild Cont. & Moderate Cont. & Heavy Cont. & Average \\

         \midrule \multirow{2}{*}{Phi-1}
         & Greedy & 0.524 & 0.421 & 0.559 & 0.728 & 0.552 \\   
         & LNE-Blocking & 0.274 (0.250) & 0.226 (0.299) & 0.169 (0.356) & 0.124 (0.400) & 0.187 (0.338)\\ 
         
        \bottomrule
    \end{tabular}
}
\end{table*}
\section{Contamination Mitigation in Small Domain-Specific Models}\label{sec:phi}
Phi-1 \cite{gunasekar2023textbooksneed} is a Transformer-based model with 1.3 billion parameters, specifically trained for fundamental Python coding tasks. Released in June 2023, Phi-1 achieves over 50\% accuracy on the HumanEval benchmark, significantly outperforming larger models such as Llama 3.1-8B (released in July 2024) \citep{dubey2024llama3}, which obtains an accuracy of 0.348. The strong performance of Phi-1 can be attributed to its training data, which consists of a carefully curated collection of "textbook-quality" web content, along with synthetically generated textbooks and exercises using GPT-3.5.

As shown in Table \ref{tab:phi}, our strategy consistently and stably restores the genuine capabilities of models with varying levels of contamination for Phi-1 on the code generation task. Notably, in comparison with Llama 2-7B and Llama 3.1-8B in Table \ref{tab:eva_code}, Phi-1 exhibits a considerably lower restored performance after applying LNE-Blocking when compared to the performance of the developer-released "uncontaminated" version of the model, with a performance gap (PG) of 25\%. This discrepancy suggests that overfitting may be a plausible concern for Phi-1, indicating potential limitations in its generalization when trained on highly curated or synthetic data.

\section{Coherence Evaluation Metrics}\label{sec:coherence}

We conducted coherence evaluations for code generation tasks using the following defined metrics.

\textbf{GPT Score (GPTS)}: We utilized the following prompts to have GPT-4o evaluate the model outputs.
\begin{itemize}
    \item  \textit{"Give you a code snippet, please rate its coherence on a scale of 1 to 10. First, provide the score enclosed in <score></score>, then give the reason enclosed in <reason></reason>. Do not pay attention to whether the code is concise or complete."  + response}
\end{itemize}
where response refers to the output generated by the model.

\textbf{Perplexity (PPL)} : We used the Meta-Llama-3.1-70B-Instruct-Turbo model to measure the perplexity of the generated content.

\textbf{Compilation Error Rate (CER)}: We calculated the ratio of SyntaxError and IndentationError occurrences during runtime over the total number of samples for the code generation task.

\section{Task-Specific Determination of Threshold\_Task}
\label{sec:threshold}
The Threshold\_Task value selected based on Llama-2 and a single task, performing effectively across all evaluated models on this task.

As shown in Table \ref{tab:determination_threshold}, It can be observed that the optimal Threshold\_Task value for the Llama-2-7B model varies across different tasks. We applied the corresponding values heuristically to other models on the same tasks and achieved stable recovery performance, as shown in Table \ref{tab:eva_code} and Table \ref{tab:eva_logical}.
\section{More Representative Generated Cases after LNE-Blocking}
\label{sec:cases}

We provide two representative examples below to illustrate the effects of applying the LNE-Blocking operation on different types of samples.

The first case involves a relatively simple task, namely generating a string of space-delimited numbers from 0 to n. After applying LNE-Blocking, the model still produces a correct solution, albeit with a different implementation style.

In contrast, the second case is more complex, requiring the model to compute the maximum nesting depth of parentheses in a given string. In this case, the memorized generation is correct, but after applying LNE-Blocking, the model's response changes to simply counting the number of left parentheses — a behavior that does not align with the intended task. This indicates that, for more challenging tasks, LNE-Blocking suppresses memorized golden answers and reveals the model's underlying, genuine reasoning capability.

\paragraph{Case 1: Simple Sequence Generation}\mbox{}\\
\textbf{Prompt:}
\begin{lstlisting}[language=Python]
def string_sequence(n: int) -> str:
    """ Return a string containing space-delimited numbers starting from 0 up to n inclusive.
    >>> string_sequence(0)
    '0'
    >>> string_sequence(5)
    '0 1 2 3 4 5'
    """
\end{lstlisting}

\textbf{Generated Content before LNE-Blocking:}
\begin{lstlisting}[language=Python]
    return ' '.join([str(x) for x in range(n + 1)])
\end{lstlisting}

\textbf{Generated Content after LNE-Blocking:}
\begin{lstlisting}[language=Python]
    retval: list[str] = []

    for i in range(n + 1):
        retval.append(str(i))

    return ' '.join(retval)
\end{lstlisting}

\paragraph{Case 2: Nested Parentheses Parsing}\mbox{}\\
\textbf{Prompt:}
\begin{lstlisting}[language=Python]
from typing import List


def parse_nested_parens(paren_string: str) -> List[int]:
    """ Input to this function is a string represented multiple groups for nested parentheses separated by spaces.
    For each of the group, output the deepest level of nesting of parentheses.
    E.g. (()()) has maximum two levels of nesting while ((())) has three.

    >>> parse_nested_parens('(()()) ((())) () ((())()())')
    [2, 3, 1, 3]
    """
\end{lstlisting}

\textbf{Generated Content before LNE-Blocking:}
\begin{lstlisting}[language=Python]
    def parse_paren_group(s):
        depth = 0
        max_depth = 0
        for c in s:
            if c == '(':
                depth += 1
                max_depth = max(depth, max_depth)
            else:
                depth -= 1

        return max_depth

    return [parse_paren_group(x) for x in paren_string.split(' ') if x]
\end{lstlisting}

\textbf{Generated Content after LNE-Blocking:}
\begin{lstlisting}[language=Python]
    import regex
    paren_groups = regex.split(r'\s+', paren_string)
    return [len(regex.findall(r'\(', p)) for p in paren_groups]
\end{lstlisting}

\begin{table*}[!htbp]
\centering
\caption{Contamination Mitigation Evaluation of Llama-2-7B on Different Datasets with Varying Values of Threshold\_Task
}
\label{tab:determination_threshold}
\resizebox{0.95\textwidth}{!}{
    \begin{tabular}{llllllll}
\toprule
Task & Model & Threshold\_Task & Uncont. & Mild Cont. & Moderate Cont. & Heavy Cont. & Average \\
\midrule \multirow{4}{*}{Code Generation} & Llama 2-7B & 0(Greedy) & 0.128 & 0.289 & 0.742 & 0.861 & 0.609 \\
  & Llama 2-7B & 1 & 0.110(0.018) & 0.171(0.043) & 0.241(0.113) & 0.262(0.134) & 0.216(0.088) \\
  & Llama 2-7B & 4 & 0.098(0.030) & 0.144(0.016) & 0.134(0.006) & 0.130(0.002) & 0.132(\textbf{0.004}) \\
  & Llama 2-7B & 7 & 0.055(0.073) & 0.091(0.037) & 0.082(0.046) & 0.077(0.051) & 0.081(0.047) \\
  \midrule
  \multirow{4}{*}{Arithmetic Reasoning} & Llama 2-7B & 0(Greedy) & 0.145 & 0.252 & 0.637 & 0.853 & 0.556 \\
  & Llama 2-7B & 1 & 0.161(0.016) & 0.200(0.055) & 0.331(0.186) & 0.379(0.234) & 0.293(0.148) \\
  & Llama 2-7B & 4 & 0.142(0.003) & 0.179(0.034) & 0.258(0.113) & 0.274(0.129) & 0.230(0.085) \\
  & Llama 2-7B & 7 & 0.133(0.012) & 0.163(0.018) & 0.224(0.079) & 0.220(0.075) & 0.198(\textbf{0.053}) \\
\bottomrule
\end{tabular}
}
\end{table*}

\section{\rebuttal{Analysis of Blocking Tokens}}

\begin{figure*}[!htbp]
\centering
  \includegraphics[width=0.9\textwidth]{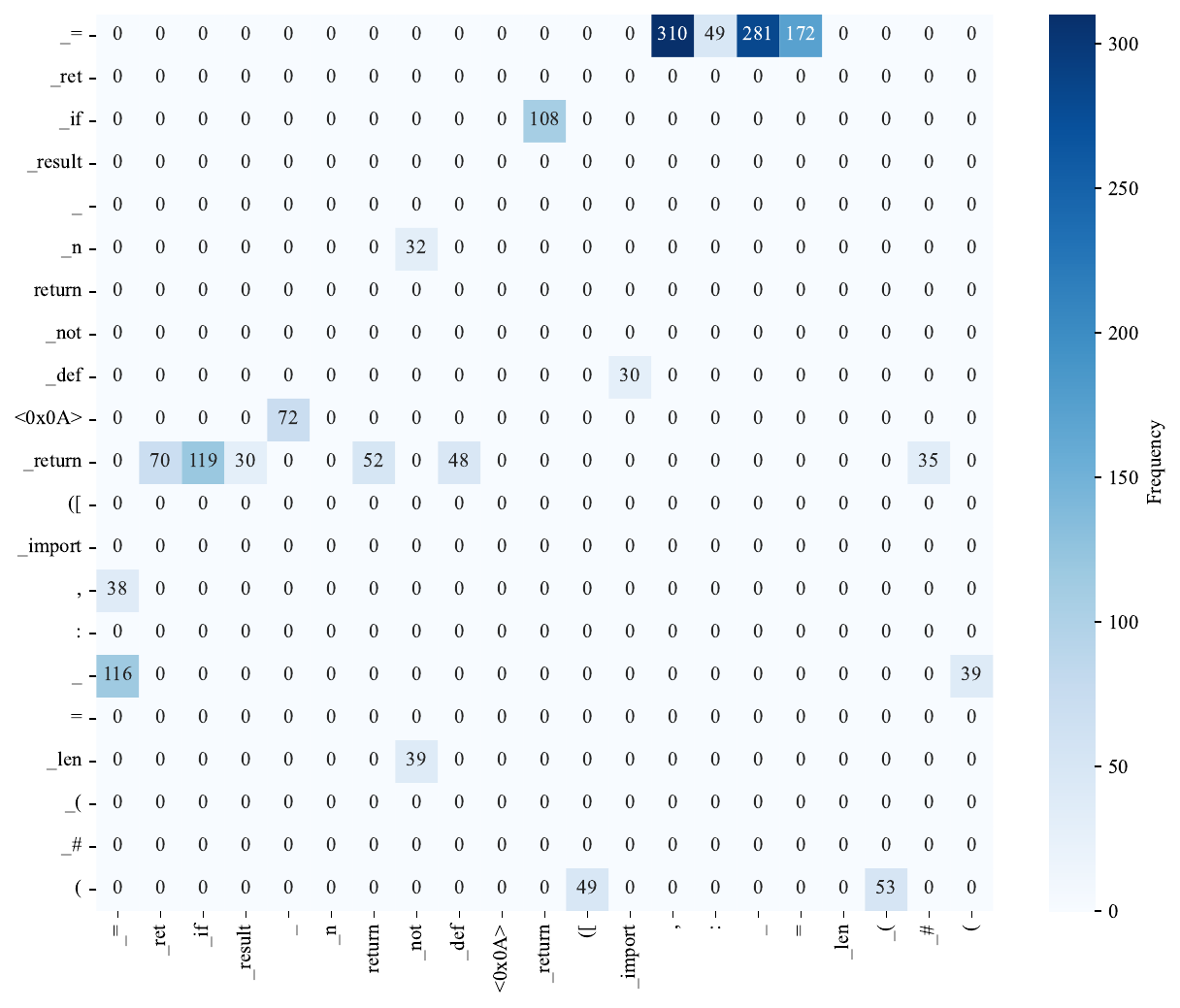}
  \caption{A heatmap of token replacement frequency during the Blocking operation.}
  
  \label{fig:heatmap}
\end{figure*}

In this section, we illustrate the token changes that occur during the execution of the Blocking operation. We pair the tokens selected for Blocking with their respective alternative tokens and visualize the top 20 tokens with the highest frequencies in Figure \ref{fig:heatmap}.

\begin{table*}[!htbp]  
\setlength\tabcolsep{6pt}
\centering
\caption{Examples of LLM Outputs Before and After Blocking, where Replaced Pairs (A, B) indicates the replacement of token A with candidate token B during Blocking. }
\label{tab:case_setence}
    \centering
    \begin{tabular}{c|p{13.5cm}}
    \toprule
        \multirow{2}{*}{ Replaced  Pairs } & Origin Outputs \\ \cline{2-2}
         & Outputs after Blocking  \\ \midrule
        \multirow{2}{*}{(\_if , \_return)} & if not numbers:$\backslash$n return 0.0$\backslash$n mean = sum(numbers) / len(numbers)$\backslash$n return sum(abs(x - mean) for x in numbers) / len(numbers)  \\
         \cline{2-2} & return float((sum(abs(x - mean(numbers)) for x in numbers)) / len(numbers))  \\ \midrule
       \multirow{2}{*}{ (\_return, \_ret)} & return [x + 1 for x in l]  \\ 
          \cline{2-2} & ret\_l=[]$\backslash$n for i in l:$\backslash$n ret\_l.append(i+1)$\backslash$n return ret\_l  \\ \midrule
        \multirow{2}{*}{(\_return, return)} & return number \% 1.0  \\ 
          \cline{2-2} & returnnumber=number\%1$\backslash$n return returnnumber  \\ \midrule
        \multirow{2}{*}{(\_return, \_if)} & return ''.join(map(lambda x, y: str(int(x) \^ int(y)), a, b))  \\ 
          \cline{2-2} & if a != b:$\backslash$n raise ValueError("Inputs must be equal")$\backslash$n return "".join(map(lambda x, y: str(int(x) \^ int(y)), a, b))  \\ \midrule
        \multirow{2}{*}{(\_return, \_\#)} & return ' '.join(str(i) for i in range(n + 1))  \\ 
          \cline{2-2} & \# TODO$\backslash$n return ' '.join(str(i) for i in range(n + 1))  \\ \midrule
        \multirow{2}{*}{(\_return, \_def) }& return [len(p) for p in paren\_string.split() if p]  \\
          \cline{2-2} & def parse(s):$\backslash$n if s[0] == '(':$\backslash$n return 1 + parse(s[1:])$\backslash$n else:$\backslash$n return 0$\backslash$n $\backslash$n return [parse(s) for s in paren\_string.split()] \\ \bottomrule
    \end{tabular}
\end{table*}

Furthermore, we provide examples in Table \ref{tab:case_setence}, displaying the complete model outputs both before and after applying the Blocking operation, to demonstrate its impact on the model's response.

\end{document}